# A multi-sensor robotic platform for ground mapping and estimation beyond the visible spectrum

Annalisa Milella[1,*], Giulio Reina[2] and Michael Nielsen[3]

[1] *Institute of Intelligent Industrial Technologies and Systems for Advanced Manufacturing, National Research Council, via Amendola 122/D, 70126, Bari, Italy*
[2] *Department of Engineering for Innovation, University of Salento, via Arnesano, 73100, Lecce, Italy*
[3] *Danish Technological Institute (DTI), Odense, Denmark*

[*]*Correspondence: annalisa.milella@stiima.cnr.it; Tel.: +39-080-592-9453*

**Abstract**

Accurate soil mapping is critical for a highly-automated agricultural vehicle to successfully accomplish important tasks including seeding, ploughing, fertilising and controlled traffic, with limited human supervision, ensuring at the same time high safety standards. In this research, a multi-sensor ground mapping and characterisation approach is proposed, whereby data coming from heterogeneous but complementary sensors, mounted on-board an unmanned rover, are combined to generate a multi-layer map of the environment and specifically of the supporting ground. The sensor suite comprises both exteroceptive and proprioceptive devices. Exteroceptive sensors include a stereo camera, a visible and near-infrared (VIS-NIR) camera and a thermal imager. Proprioceptive data consist of the vertical acceleration of the vehicle sprung mass as acquired by an inertial measurement unit (IMU). The paper details the steps for the integration of the different sensor data into a unique multi-layer map and discusses a set of exteroceptive and proprioceptive features for soil characterisation and change detection. Experimental results obtained with an all-terrain vehicle operating on different ground surfaces are presented. It is shown that the proposed technologies could be potentially used to develop all-terrain self-driving systems in agriculture. In addition, multi-modal soil maps could be useful to feed farm management systems that would present to the user various soil layers incorporating colour, geometric, spectral and mechanical properties.

**Introduction**

Future agricultural vehicles will be required to operate throughout extensive fields, with limited or no human supervision, while preserving driving safety. One of the challenges in this context is the ability to automatically build accurate soil maps, as well as to perceive and analyse the properties of the traversed ground (Ball et al. 2015). Soil mapping and estimation are beneficial for a vehicle to cope with its environment more efficiently and to better support its tasks. The knowledge of ground characteristics also contributes to increase the safety of vehicles during operations near ditches or on hillsides and cross slopes, and on hazardous highly-deformable terrain. At the same time, high-resolution soil maps are necessary to achieve site-specific operation management in precision farming.

In most agricultural applications, a coarse map of the soil where the vehicle operates is typically available. Such maps are often created off-line, and then loaded into the navigation control system. However, they may contain errors due to recent changes in the field caused by humans or by nature. Furthermore, they do not provide any information about moving obstacles (e.g., human beings, animals and vehicles), nor about the nature of the ground to be traversed. This causes a major safety issue related to any type of semi- or fully-autonomous navigation and operation.

As a result, a large body of research has been recently devoted to develop new strategies to increase the perception capabilities of agricultural robots (Ross et al. 2015; Reina and Milella 2012), mainly using vision sensors (Rovira-Más et al. 2008; Dong et al. 2017), radar (Milella et al. 2015), and lidar (Kragh et al. 2015). Alternatively, approaches based on RGB-depth (RGB-D) cameras have been proposed by Marinello et al. (2015) to characterise soil during tillage operations, and by Nissimov et al. (2015) to develop a navigation system for greenhouse operations. However, the selection and integration of the most appropriate sensors to generate detailed and highly informative soil maps, remains an on-going research topic.

This paper extends previous research by the authors (Milella et al. 2017) in the development and implementation of advanced ground mapping and estimation approaches, based on multi-sensor platforms and data processing algorithms, to be integrated onboard off-road vehicles, including agricultural machinery[1]. The aim of this research is to facilitate operations on a narrow scale with a smaller environmental footprint that may be useful to increase precision (e.g., in farming applications), to provide fast automated safety responses or traversability assessment, and for controlled traffic purposes in general.

Specifically, a modular sensing box is proposed, which integrates complementary sensors to generate ground maps including 3D geometric and spectral (in the visible and beyond) features, as well as vibration response information.

The idea of combining heterogeneous optical sensors for environment modelling has been little investigated in the past. For instance, in Nieto et al. (2010) the fusion of laser and hyperspectral data was proposed to generate 3D geological models of the environment. In Rangel et al. (2014), thermal, visual and depth information obtained from a thermography camera and a Microsoft Kinect depth camera were integrated to obtain 3D-thermal models for inspection tasks. The use of imaging spectroscopy and thermal imaging also in combination with LIDAR and stereo cameras has been investigated for remote soil monitoring and mapping (Mulder et al. 2011).

The use of such sensing techniques on a robotic vehicle for sensing the surrounding environment at a local level represents the novel and original contribution of this work.

As a further contribution, the integration of proprioceptive information[2] is proposed in the form of vehicle vertical motion induced by terrain irregularities as an additional map layer. While exteroceptive sensors can provide useful information about the terrain appearance and status, important complementary information can be drawn from proprioceptive sensors, as the latter are able to measure physically-based dynamic effects that govern the vehicle-terrain interaction and that greatly affect its mobility. Recent research has focused on terrain estimation via proprioceptors. For example, vibrations induced by wheel-ground interaction were fed to different types of terrain classifiers that discriminated based on, respectively, Mahalanobis distance of the power spectral densities (Brooks and Iagnemma 2005), AdaBoost (Krebs et al. 2009), neural network (Dupont et al. 2008) and Support Vector Machine (SVM) (Reina et al. 2017). Here, a method to associate acceleration measures from an IMU to a 3D visual map is developed, in order to enrich the map with information related to the vehicle vibration response to the excitation induced by terrain irregularity. Specifically, a given ground region is, first, reconstructed from a distance using visual data, then, when the same region is traversed by the vehicle, proprioceptive measurements are associated with it, based on the knowledge of vehicle displacement via a visual odometry algorithm. The resulting multi-modal maps are successively processed to extract statistical features to characterise the traversed ground and detect significant ground changes based on a cumulative sum (CUSUM) test (Page, 1954).

---

[1] FP7ERA-NET ICT-AGRI2 Simultaneous safety and surveying for collaborative agricultural vehicles (S3-CAV; http://s3cav.eu/).

[2] In the context of this research, proprioceptive sensors measure values internally to the robot, e.g. vertical accelerations. Exteroceptive sensors are used for the observation of the environment, e.g. colour and geometry of the ground.

Experimental results obtained with an all-terrain vehicle operating on different surfaces are presented to validate the proposed framework. It is shown that the proposed technologies could be potentially adopted to increase vehicle mobility and safety, as well as for precision agriculture applications.

## Materials and Methods

### System Overview

A modular sensing box, to be integrated onboard ground vehicles, is presented that incorporates complementary sensors to generate 3D ground maps including geometric and spectral (in the visible and beyond) features, as well as vibration response information. The sensor suite comprises two visual cameras forming a stereo couple, a VIS-NIR sensor, a thermal camera, and an inertial measurement unit (IMU). The idea is that, by registering and integrating all data, an informative multi-modal representation of the ground can be built.

In summary, the proposed framework proceeds as follows. First, for each incoming video frame, a stereo reconstruction algorithm is applied to recover a 3D representation with colour content of the scene at the current time instant. Then, the stereo-generated point cloud is associated with thermal and VIS-NIR data, leading to 3D thermal and 3D VIS-NIR environment representations, respectively.

The integration of different sensor information is beneficial in many ways. For instance, stereo reconstruction may help in differentiating between ground and obstacles, by adding height information, whereas the use of thermal data may simplify the identification of thin objects or living beings thanks to their different emissivity and temperature. The spectral signature returned by the VIS-NIR sensor may help to discriminate different ground surfaces, as well as to get information on the vegetation status through appropriate metrics.

In addition, as the vehicle moves across its operating space and senses the supporting plane through its proprioceptive sensors, the ground map is also enriched with information pertaining to the vehicle-terrain interaction. In particular, acceleration data are stored and associated with robot positions in the map.

It is important to note that data superposition is only possible when the field of view of the different sensors overlap. This aspect poses many challenges since sensors differ in the extent of the field of view and sampling rate. Therefore, it is very important to register each sensor scan with the corresponding robot position. In this research, this is achieved by resorting to a visual odometry algorithm.

A feature-based characterisation of the traversed terrain in terms of appearance, spectral and "mechanical" properties is also proposed. For this aim, the ground map is segmented into samples or patches, which are processed to extract a set of statistical features using the different data sources. The extracted features are then used to feed a change detection algorithm, which allows detection of terrain changes along the vehicle path.

Details about the system setup and the developed algorithms are provided in the following.

**Experimental test bed**

The experimental test bed employed for sensor data collection is shown in Figure 1(a). The robotic base features a four-wheel drive system with skid-steering. The perception system is composed of a self-contained sensor box custom-built at the Danish Technological Institute (DTI). The sensor box is visible attached to the robot's sensor frame (Figure 1(a)) and in an off-the-shelf picture (Figure 1(b)). It comprises: two Basler DART DaA1600-60uc modules (Basler AG, Ahrensburg, Germany) for stereo vision with 40 mm baseline; one thermal camera Micro Epsilon ThermalImager160 (Micro-Epsilon, Ortenburg, Germany); one hyperspectral VIS-NIR Specim V10 on a Basler ACA1920-155um 400-1000nm (Basler AG, Ahrensburg, Germany).

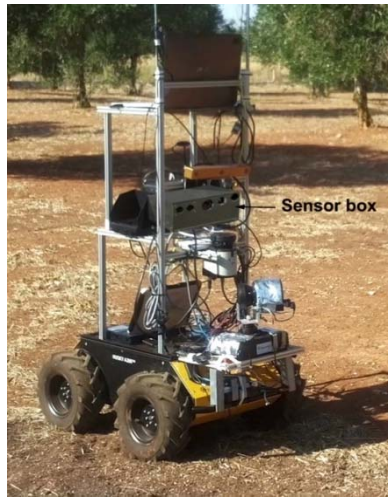

(a)

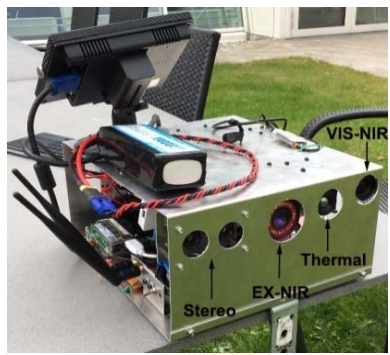

(b)

**Figure 1** (a) The all-terrain vehicle used for field validation. (b) Off-the-shelf sensor box.

The hyperspectral camera is a line-scan system, therefore the spatial dimension is collected through vehicle movement. The stereo camera delivers up to 1600×1200 pixels at 60 Hz, which was reduced to 7.5 Hz in the experiments to decrease storage capacity requirements. The thermal imager can deliver 160×120 pixels at 120 Hz, reduced to 15 Hz. The VIS-NIR camera setup was 1920 pixel lines with 2 nm digital resolution at 128 Hz, optically limited to 8 nm usable bands. 2 nm actual resolution would require the significantly larger V10E spectrograph. The thermal camera and the hyperspectral camera grab 16-bit images, while the colour cameras are 8-bit.

Furthermore, the sensor box features a x-IMU (x-io Technologies Limited, Colorado Springs, USA) set to run at 128 Hz. It is commonly referred to as a 9DoF IMU, that is, it comprises triple axis gyroscope, triple axis accelerometer and triple axis magnetometer to provide a drift-less measurement of orientation relative to the Earth reference system.

Finally, the box also contains DC/DC conversion from 9-36 V to 12 V. It has under-voltage protection and voltage balancing between two inputs, making it safe and easy to hotswap between two lithium polymer (LiPo) batteries. The embedded computer is an Advantech MIO 5271 I5 dual core processor (Advantech Co., Taipei, Taiwan) running Windows 10 IoT, and two system fans. The weight of the system without battery is 5 kg, and it fits in a carry-on luggage with its 0.32×0.28×0.14 m dimensions.

**Sensor synchronization and calibration**

The association of heterogeneous data requires time synchronization. A timestamp-based synchronization approach was adopted, whereby each sensor observation was marked with a timestamp. In addition, to register all sensor data with respect to a common reference frame, a calibration phase is required to estimate the intrinsic and extrinsic parameters of the sensors. Calibration was performed only once after the assembly of the device and comprised the following steps.

- *Calibration of stereo device*: the stereo pair was calibrated using the Camera Calibration Toolbox for MATLAB (Bouguet 2008). Both intrinsic (i.e., focal length, principal point, radial and tangential distortion coefficients) and extrinsic parameters (i.e., the relative position and orientation of the cameras with respect to each other) were estimated based on a set of images of a planar checkerboard that was appropriately moved across the field of view of each camera of the stereo device (Figure 2(a)). The calibration functions also returned the rectification matrices to rectify the images as a preliminary step before applying the stereo-matching algorithm. An additional calibration step was successively performed to estimate the position and orientation of the stereo device with respect to a reference frame attached to the vehicle. To this aim, the calibration pattern was positioned at a known location with respect to the vehicle. Then, the extrinsic parameters relative to the pattern and, consequently, the position and orientation relative to the vehicle were inferred, using a least-squares optimisation process.

- *Calibration of thermal camera*: the procedure employed for the calibration of the visual cameras was adapted to estimate the intrinsic parameters of the thermal device. In particular, to make corners visible, the checkerboard was first exposed under a heat lamp.

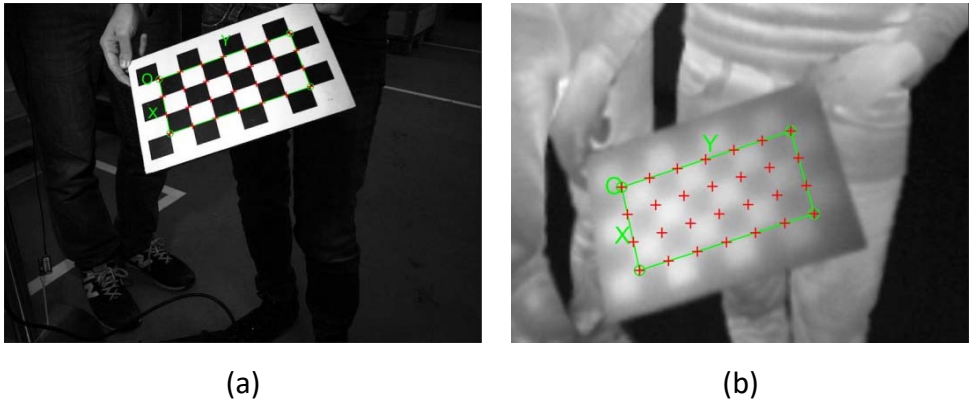

(a)          (b)

**Figure 2** Sample images acquired for calibration of (a) visual and (b) thermal cameras.

- *Calibration of thermal with respect to stereo camera*: this calibration step was aimed at estimating the position and orientation of the thermal camera with respect to the stereo device. In particular, the transformation matrix relating the position and orientation of a reference frame attached to the thermal camera and a reference frame attached to the left camera of the stereo system was estimated. To this end, the calibration pattern was positioned at different locations in the environment, and images of the pattern were simultaneously acquired by the left camera of the stereo device and by the thermal camera. Sample calibration images are shown in Figure 2. Then, the extrinsic parameters of each camera relative to the pattern and, consequently, the position and orientation of the two cameras with respect to each other were inferred using a least-squares optimisation process. Such a relationship is the basis for the back-projection of thermal image data onto the stereo reconstructed map, as will be described in the next section.

- *Calibration of VIS-NIR with respect to stereo camera*: this calibration step was aimed at finding the position of the scan line of the VIS-NIR sensor in the visual image. A fluorescent lamp was employed for this calibration task, as shown in Figure 3. The position of the scan line was found by moving the lamp across the floor so as to match the full field of view, i.e., the scan line, of the VIS-NIR camera. The aim was to find the line in the RGB that the VIS-NIR overlaps, in order to later map the VIS-NIR on top of the stereo image at that line.

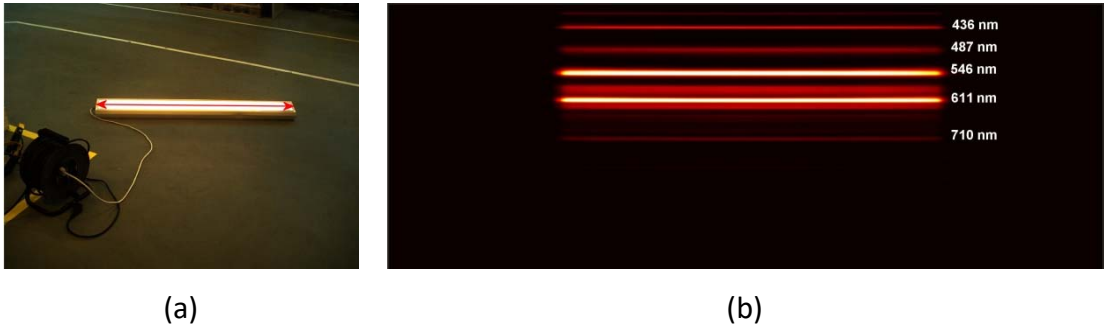

(a)            (b)

**Figure 3** Sample images acquired during calibration of the VIS-NIR with respect to the visual sensor: (a) visual image with the position of the VIS-NIR scan line overlaid; (b) VIS-NIR image with overlaid wavelength information.

- *Calibration of the IMU*: The x-IMU provides as outputs the acceleration readings, $a = [a_x, a_y, a_z]$ measured in the sensor reference frame and the quaternion, *q* that expresses the orientation of the sensor reference frame with respect to the Earth reference frame. In order to get the dynamic accelerations, the vector gravity $g = [0, 0, -9.81\frac{m}{s^2}]$ needs to be compensated for. Therefore, first the accelerometer readings *a* are rotated in the Earth frame according to Eq. 1:

$$a_E = q \begin{pmatrix} 0 \\ a \end{pmatrix} q^{-1} \qquad (1)$$

then, the dynamic acceleration, $a_E^d$ can be obtained by subtracting gravity following Eq. 2:

$$a_E^d = a_E - g \qquad (2)$$

**Multi-modal mapping**

In this section, the multi-modal mapping approach is described in detail. First, the stereo-based mapping algorithm is introduced. Then, the generation of a 3D map enriched with thermal and VIS-NIR information is presented. Finally, the integration of acceleration data with visual information is discussed.

*Stereo-based mapping*

Stereo data provides colour and geometric information that can be used to generate a 3D map of the environment. For each stereo pair, the stereo processing algorithm includes the following steps:

- *Rectification*: each image plane is transformed so that pairs of conjugate epipolar lines become collinear and parallel to the horizontal image axis; this reduces the problem of computing correspondences from a 2D to a 1D search problem.
- *Disparity map computation*: to compute the disparity map, the Semi-Global Matching (SGM) algorithm is used. This algorithm combines concepts of local and global stereo methods for accurate, pixel-wise matching with low runtime (Hirschmuller 2005).
- *3D point cloud generation*: as the stereo pair was calibrated both intrinsically and extrinsically, disparity values can be converted to depth values and 3D co-ordinates can be computed in the reference camera frame for all matched points. Point co-ordinates can also be successively transformed from camera frame to vehicle frame, using the transformation matrix resulting from the calibration process. A statistical filter is applied to reduce noise and remove outlying points. A voxelised grid approach is finally used to reduce the number of points and decrease the computational burden.

Subsequent point clouds are stitched together to produce a 3D map. To this end, the vehicle displacement between two consecutive frames is estimated using the open source visual odometry algorithm *libviso2* (Geiger et al. 2011). A box grid filter approach is then applied for point cloud merging. Specifically, first, the bounding box of the overlapping region between two subsequent point clouds is computed. Then, the bounding box is divided into cells of specified size, and points within each cell are merged by averaging their locations and colours. Points outside the overlapping region remain untouched.

*Integration of stereo and thermal data*

The corresponding thermal 3D point cloud is obtained as follows. First, knowing the position and orientation of the thermal sensor with respect to the stereo device, the 3D co-ordinates of the points pertaining to each stereo reconstructed point cloud are transformed from the stereo to the thermal reference frame. Then, knowing the intrinsic parameters of the thermal camera, 3D points are projected in the thermal image, and their thermal values are extracted and assigned to the corresponding 3D points in the stereo map. The result of stereo-thermal integration for a sample image is shown in Figure 4, for grass soil with a manhole cover.

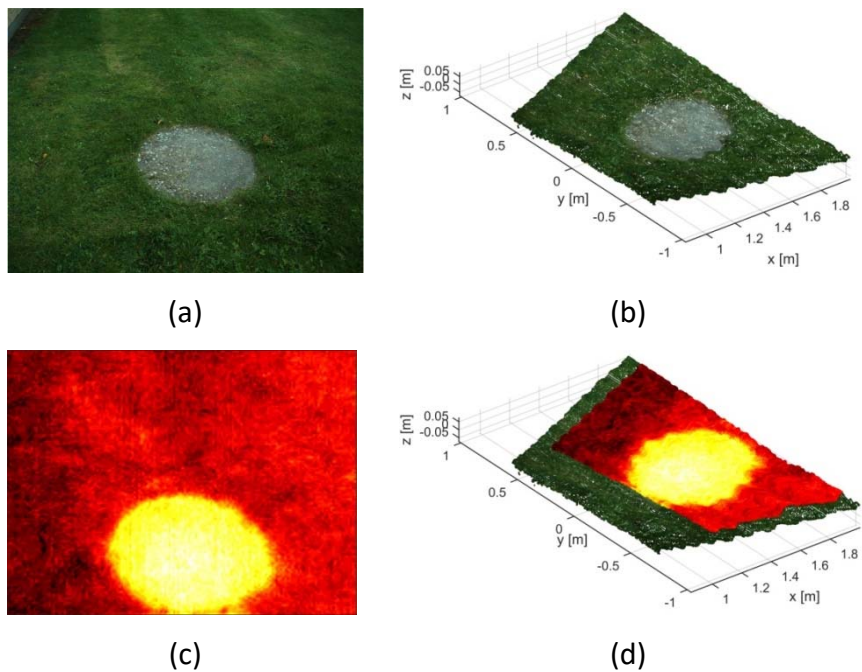

(a)          (b)

(c)          (d)

**Figure 4** Thermal and stereo integration for a sample test case (a manhole cover): (a) visual image from left camera of the stereo system; (b) stereo reconstructed 3D point cloud; (c) thermal image; (d) thermal data overlaid on the stereo 3D point cloud. In (c) and (d), brighter colours correspond to regions with higher temperature.

Specifically, Figure 4(a) shows the visual image acquired by the left camera of the stereo device, while Figure 4(b) displays the stereo reconstructed point cloud. The synchronised thermal image is reported in Figure 4(c). Finally, the stereo point cloud with the overlaid thermal layer is shown in Figure 4(d). The latter provides a multi-modal 3D representation of the terrain, with stereovision information representing geometric and visual appearance properties of the ground, and thermal data providing temperature information. It can be noticed that the manhole cover contrasts better in the thermal image than in the visual one, therefore, in this case, thermal data help discriminating different parts of the scene.

As a further example, Figure 5 shows a test image acquired in a forest environment. A thin trunk lying on the ground is clearly visible when adding thermal information, whereas it is not easily detectable using visual data only, due to low contrast and critical illumination conditions.

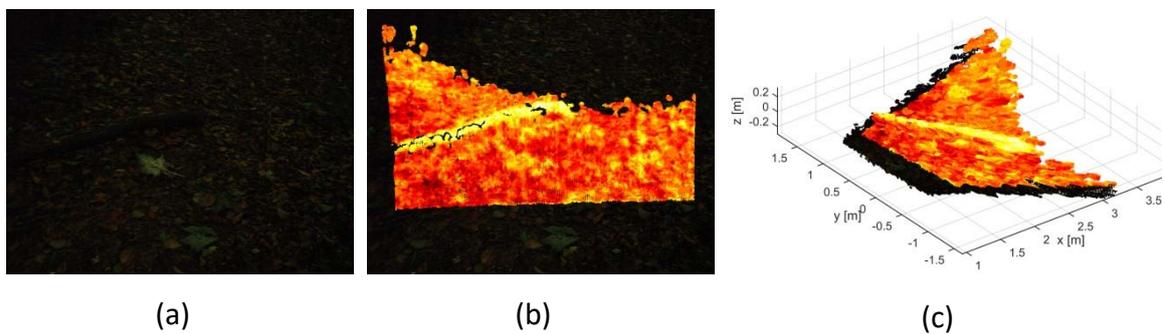

(a) (b) (c)

**Figure 5** Integration of thermal and stereo data for a sample image (a forest scene): (a) visual image; (b) visual image with overlaid thermal data; (c) thermal data overlaid on the stereo 3D point cloud.

*Integration of stereo and VIS-NIR data*

A 3D environment representation augmented with VIS-NIR information is obtained by integrating each stereo-reconstructed point cloud with VIS-NIR data. In detail, based on calibration information, the scan line of the VIS-NIR sensor is projected onto the image acquired by the reference (left) camera of the stereo device. Then, the corresponding 3D points reconstructed by stereovision are extracted, and they are assigned the spectral signature returned by the VIS-NIR sensor. An example is shown in Figure 6. In particular, Figure 6(a) displays the visual image with the VIS-NIR scan line overlaid that is located at approximately 2 m ahead of the vehicle. The corresponding three-dimensional reconstruction is shown in Figure 6(b). The VIS-NIR image is displayed in Figure 6(c), where the light wavelength content and the horizontal scan distance are shown respectively along the vertical and horizontal axis. Note that, for visualization purposes, in Figure 6(a) and (b) each point of the projected VIS-NIR scan line is assigned a false colour where the red channel value is given by the average of the first 100 bands of the VIS-NIR image, the green channel value is given by the average of the following 100 bands, and the blue channel value corresponds to the average of the last 100 bands. Finally, using motion information provided by stereo visual odometry, subsequent scan lines can be stitched together, thus producing an additional layer, i.e., the VIS-NIR layer, to the 3D map. An example is shown in Figure 7, for a test comprising two different terrains (i.e., grass and paved road). Specifically, Figure 7 (a) shows the 3D stereo map while, in Figure 7 (b), its corresponding VIS-NIR layer is displayed with false colours. It can be seen that the VIS-NIR layer may help to discriminate parts of the scene such as the light pole, which shows lower contrast in the visual layer than in the VIS-NIR one.

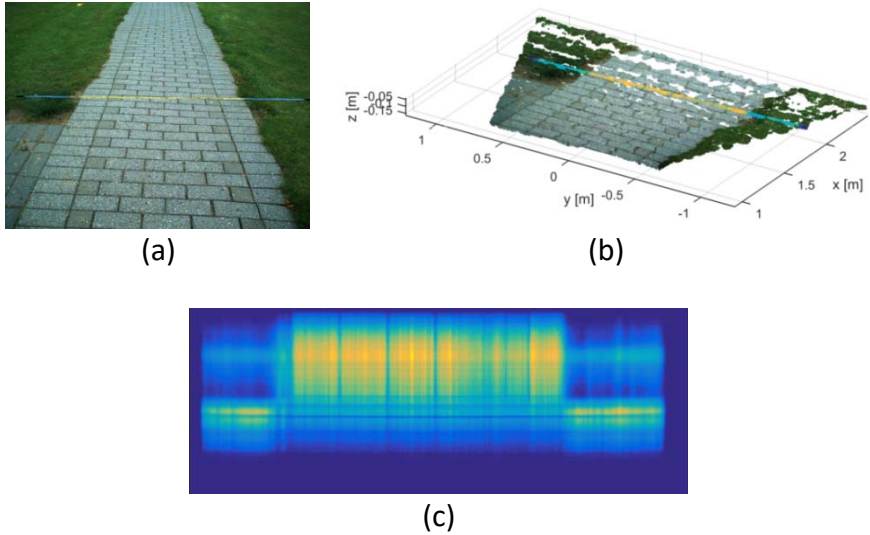

(a)                          (b)

(c)

**Figure 6** Integration of stereo and VIS-NIR data. (a) Visual image with VIS-NIR scan line overlaid. (b) 3D stereo reconstruction with VIS-NIR scan line overlaid. (c) VIS-NIR image.

The VIS-NIR layer can be also used to extract metrics of interest, such as a suitable index. In this work, the normalized difference vegetation index (NDVI) is considered, as one of the most commonly used vegetation indices. The NDVI provides a measure of healthy, green vegetation. Eq. 3 shows its calculation:

$$NDVI = \frac{NIR-RED}{NIR+RED} \quad (3)$$

where *RED* and *NIR* stand for the spectral reflectance measurements acquired in the red (visible) and near-infrared regions, respectively. The NDVI ranges between -1 and

1. Healthy plants produce high NDVI (0.6 or greater) and in the other case, the results are between 0 and 0.3. Lifeless zones produce low results (lower or equal to 0). Centre *RED* and *NIR* wavelengths for the calculation of the NDVI were set at 670 nm and 800 nm respectively, as suggested in the literature (Wu et al. 2008; Underwood et al. 2017).

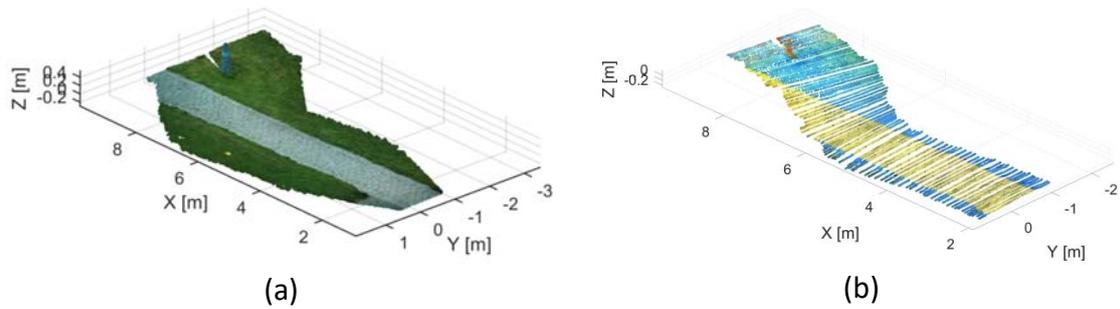

(a)  (b)

**Figure 7** (a) Visual and (b) VIS-NIR layer (displayed in false colours) of a 3D map obtained for a test including two different types of terrain (i.e., grass and paved road).

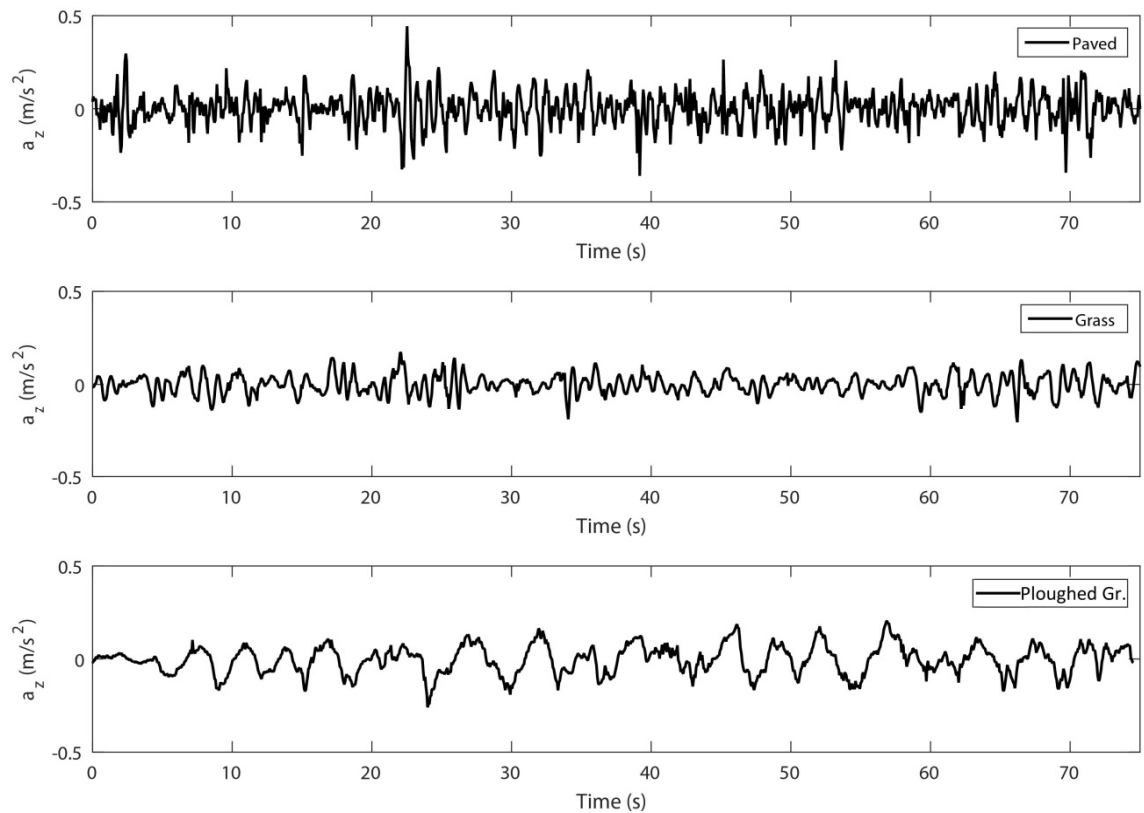

**Figure 8** Vertical accelerations experienced by the vehicle on different soil surfaces: (top) Stone-paved path, (middle) Grass, (bottom) Ploughed ground.

*Integration of stereo and acceleration data*

Additional information reflecting the vibration response of the robot is added to the 3D visual map by using acceleration data acquired by the IMU. Sample results are shown in Figure 8 where the vehicle vertical accelerations induced by three different surfaces (paved road, grass and ploughed ground) are considered. In all experiments, the robot was commanded to follow a straight line at constant speed of about 0.8 m/s. As expected, the harder the soil surface, the higher the vibration level with a root mean square (RMS) value for $a_z$ of about 0.050, 0.065 and 0.085 m/s$^2$, respectively for grass, ploughed ground and paved road.

It is important to note that, due to its forward-looking configuration, the stereo camera surveys the environment in front of the vehicle, whereas proprioceptors measure the current supporting surface. Therefore, a given terrain portion is, first, reconstructed from a distance using stereovision information, then, when the same region is traversed by the vehicle, proprioceptive measurements can be associated with it. The match of the visual and proprioceptive data requires the estimation of the successive displacements of the vehicle, which is performed via visual odometry. The proper match between exteroceptive and proprioceptive data is also important for the successive feature extraction phase, as will be described in the following section.

**Multi-modal ground estimation**

A multi-modal ground estimation approach is proposed, whereby as the vehicle proceeds along its path, ground patches (or samples) are characterised by a set of features extracted from the different data sources. A ground change detection algorithm is also developed to detect significant variation in the ground characteristics, which may affect vehicle mobility or other operations in the field.

In the context of this research, a ground patch results from the segmentation of the stereo-generated three-dimensional reconstruction using a window-based approach. First, only the points that fall below the vehicle's body or undercarriage are retained, thus discarding parts of the environment not directly pertaining to the ground such as bushes, trees and obstacles in general. Then, ground patches are formed incorporating (i.e., stitching) data acquired in successive acquisitions using vision-based localisation.

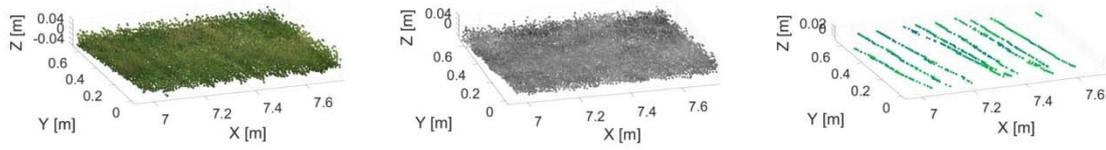

**Figure 9** Ground patch in the visual (a), thermal (b) and VIS-NIR (c) layers for a test on grass. In the thermal layer, brighter points correspond to points with higher thermal radiance, while the VIS-NIR layer displays NDVI values ranging from -1 (blue) to 1 (green).

In the current implementation, ground patches were obtained by stitching four consecutive frames. Given the 7.5 Hz acquisition frame rate of the stereo device and the 128 Hz acquisition rate of the IMU device, an average vehicle operating speed of 0.8 m/s and the dimensions of the vehicle's chassis (0.54(L) ×0.70(W) m), terrain patches of approximately 0.85×0.7 m with a corresponding acceleration signal of about 50 readings were considered for feature extraction. As an example, Figure 9 shows a ground patch for a test on grass in the visual (Figure 9(a)), thermal (Figure 9(b)) and VIS-NIR (NDVI) (Figure 9(c)) layers.

### *Description of the feature sets*

### *Colour features*

A stereo camera typically provides raw colour data as red, green and blue (RGB) intensities. However, this representation suffers from sensitivity to changes in lighting conditions, possibly leading to poor classification results. To mitigate this issue, the so-called $c_1c_2c_3$ colour model is used (Gevers and Smeulders 1999). Eqs. 4-6 illustrate its derivation,

$$c_1 = \tan^{-1}\left(\frac{R}{\max(G,B)}\right) \quad (4)$$

$$c_2 = \tan^{-1}\left(\frac{G}{\max(R,B)}\right) \quad (5)$$

$$c_3 = \tan^{-1}\left(\frac{B}{\max(R,G)}\right) \quad (6)$$

where *R*, *G* and *B* are the pixel values in the RGB space.

For each ground patch, colour features were obtained as statistical measures extracted from each channel *i* of the $c_1c_2c_3$ space according to Eqs. 7-10:

$$E_i = \frac{1}{N}\sum_{n=1}^{N} x_n \quad (7)$$

$$\sigma_i^2 = \frac{1}{N}\sum_{n=1}^{N}(x_n - E_i)^2 \tag{8}$$

$$Sk_i = \frac{\frac{1}{N}\sum_{n=1}^{N}(x_n - E_i)^3}{\left(\sqrt{\frac{1}{N}\sum_{n=1}^{N}(x_n - E_i)^2}\right)^3} \tag{9}$$

$$Ku_i = \frac{\frac{1}{N}\sum_{n=1}^{N}(x_n - E_i)^4}{\left(\sqrt{\frac{1}{N}\sum_{n=1}^{N}(x_n - E_i)^2}\right)^4} \tag{10}$$

where $x_n$ is the intensity value associated with a point in one of the three colour channels, and $N$ refers to the total number of points in the patch. The mean $E_i$ (Eq. (7)) defines where the colour lies in the $c_1c_2c_3$ colour space. The variance $\sigma_i^2$ (Eq. (8)), indicates the spread or scale of the colour distribution. The skewness $Sk_i$ (Eq. (9)) provides a measure of the asymmetry of the data around the sample mean. The Kurtosis $Ku_i$ (Eq. (10)) measures the flatness or peakedness of the colour distribution. Thus, the colour properties are presented by four features for each channel for a total of twelve elements.

*Thermal features*

As for colour data, mean ($E_{th}$), variance ($\sigma_{th}^2$), skewness ($Sk_{th}$) and kurtosis ($Ku_{th}$) can be extracted from thermal data associated with the points belonging to a given ground patch and used as thermal features. Given a histogram of pixel intensities from a thermal image, statistical features are related to the probability of observing a gray-level value at a random location in the sample (Fehlman and Hinders 2009). In detail, $E_{th}$ and $\sigma_{th}^2$ denote respectively mean and variance of the radiance from the ground and any emitting objects therein. $Sk_{th}$ is the skewness and $Ku_{th}$ the kurtosis, measuring respectively the degree of histogram asymmetry around the mean and the histogram sharpness.

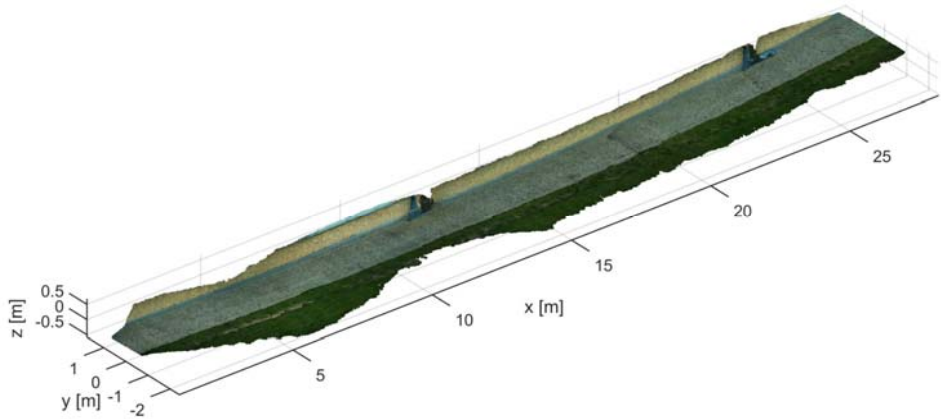

**Figure 10** 3D stereo-generated map obtained for a test on paved road. The environment includes paved road, grass, a lateral wall, and some small obstacles (i.e., poles).

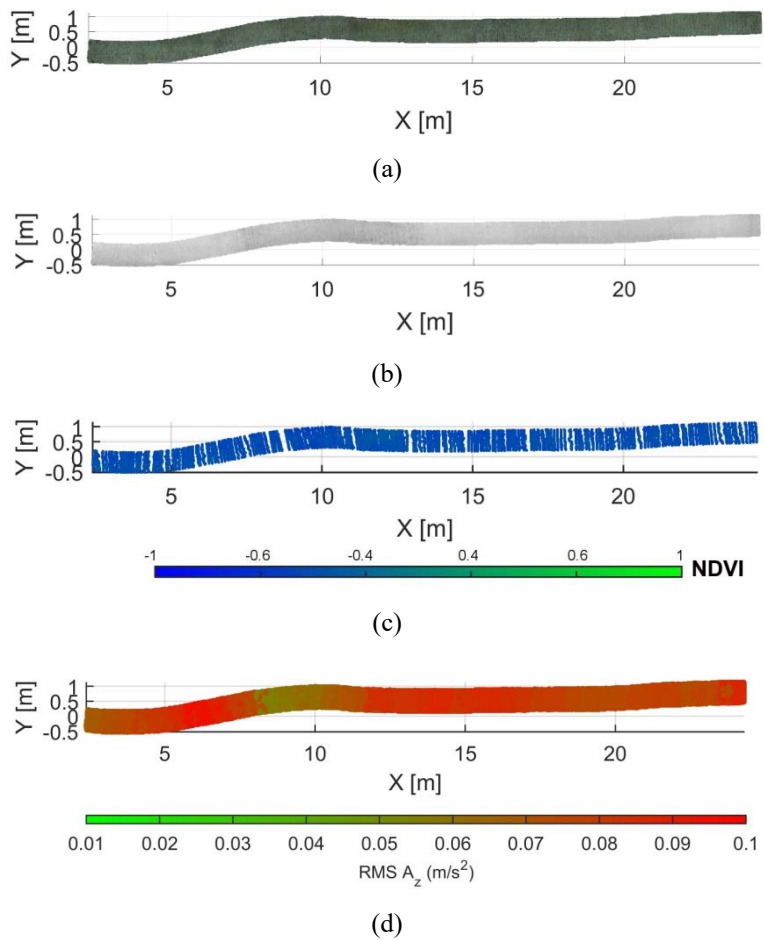

**Figure 11** Multi-layer map of the traversed ground (i.e., paved road) for the test of Fig. 10: top view of (a) visual (RGB), (b) thermal, (c) VIS-NIR, and (d) vibration layer.

*VIS-NIR features*

Similar to what is done for colour and thermal features, in this analysis, VIS-NIR features are defined as mean ($E_v$), variance ($\sigma_v^2$), skewness ($Sk_v$) and kurtosis ($Ku_v$) of NDVI values associated with the points in a given ground patch.

*Acceleration features*

Proprioceptive features can be extracted from acceleration data registered during the traversal of a given ground patch. In particular, the root mean square (RMS) of the vertical acceleration is used, according to Eq. 11,

$$RMSa_z = \sqrt{\frac{1}{M}\sum_{i=1}^{M} a_{z_i}^2} \qquad (11)$$

where $i = 1, 2,...M$ denotes the $i^{th}$ acceleration reading and $M$ is the number of acceleration readings in the window considered. This feature provides an estimate of the magnitude of the vertical vibrations experienced by the vehicle.

*Change detection algorithm*

Significant variations in the ground properties during normal vehicle operation can be detected resorting to the CUSUM test, which is based on the cumulative sums charts. The CUSUM test is computationally very simple and intuitive, and it is fairly robust to different types of changes (abrupt or incipient). In words, the CUSUM test looks at the prediction errors $\epsilon_t$ of the value of a given feature. Under the assumption of normally distributed data $\epsilon_t = (x_t - \bar{x}_t)/\sigma$, where $x_t$ is the last point monitored, $\bar{x}_t$ the mean of the process before the observation $x_t$ and $\sigma$ its standard deviation. $\epsilon_t$ is a measure of the deviation of the observation from the target: the farther the observation is away from the target, the larger $\epsilon_t$. The CUSUM test gives an alarm when the recent prediction errors have been sufficiently positive for a while.

## Results and discussion

This section reports the field validation of the proposed framework for multi-modal ground mapping and estimation. Tests were performed by driving the vehicle on different ground types as the on-board sensors acquired data from the surrounding environment. Then, the proposed algorithms were applied offline.

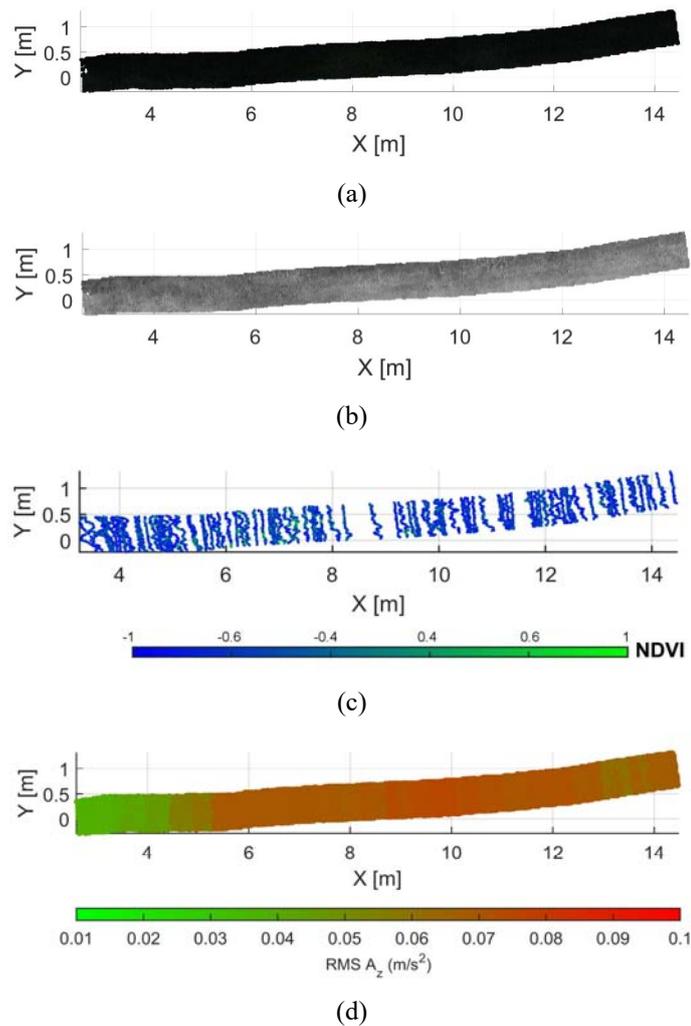

**Figure 12** Multi-layer map obtained for a test on ploughed ground: top view of (a) visual (RGB), (b) thermal, (c) VIS-NIR, and (d) vibration layer.

In the following, first, results of the integration of stereo data respectively with thermal, VIS-NIR and IMU data to build multi-layer soil maps are reported. Then, the feature sets are used for ground change detection.

**Ground mapping**

Multi-modal ground maps were constructed with the vehicle moving on different ground types. Specifically, stone-paved, ploughed ground and grass were considered. Figure 10 shows the whole map generated by stereo-vision for a test with the vehicle moving on a stone-paved path of about 25 m. The test environment also includes grass, a lateral wall and some small obstacles (i.e., poles). The top view of the corresponding spectral map referring to the traversed ground is shown from Figure 11(a) to 11(c), respectively, in the visible, thermal and VIS-NIR spectrum. For the same path, the ground "signature" expressed in terms of RMS accelerations along the Z-axis is displayed in Figure 11(d). As shown in these figures, the stone-paved surface features a relatively large thermal emissivity and a low (less than 0) NDVI coefficient. The vibration level is relatively high due to the surface hardness. Similarly, Figure 12 refers to the case of the vehicle traversing ploughed ground. In this case, the thermal and vibration signatures decrease, while the NDVI index still results in negative values due to the absence of vegetation. Finally, Figure 13 reports the multi-layer map constructed for a path on grass, where the NDVI index is higher (close to 0.4) and the vibration response is modulated by the surface softness.

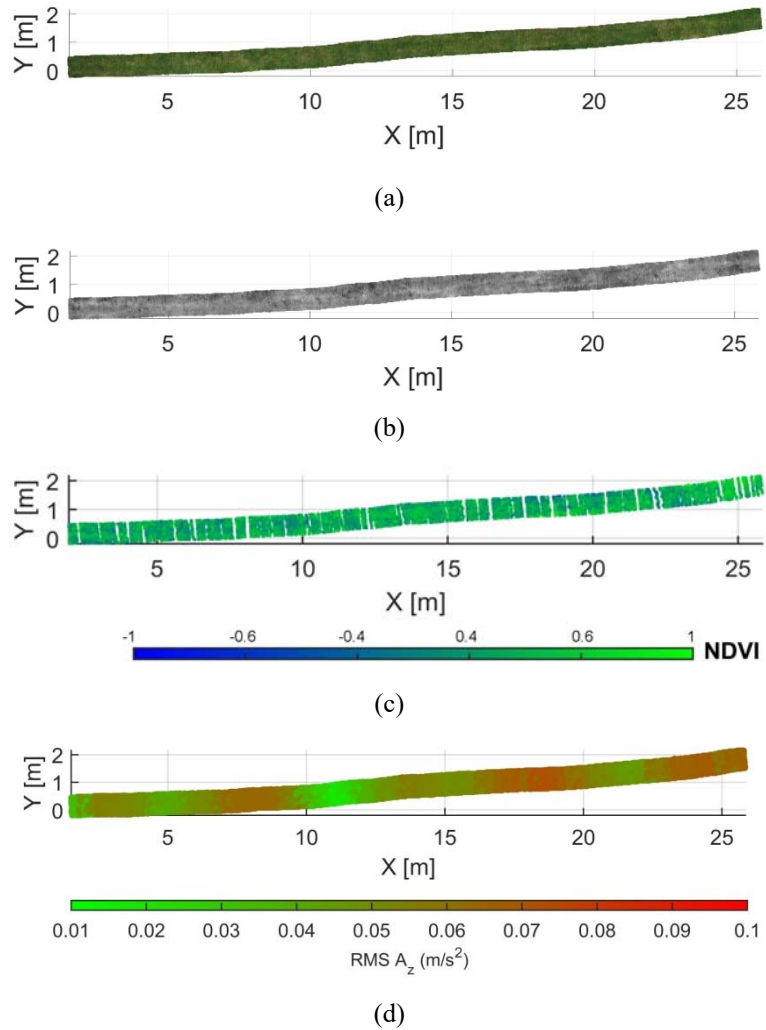

**Figure 13** Multi-layer map obtained for a test on grass: top view of (a) visual (RGB), (b) thermal, (c) VIS-NIR, and (d) vibration layer.

It can be seen that the integration of the different sensor types leads to highly informative maps, which can be potentially used to support any precision application that relies on the accuracy of ground knowledge and representation at local level, such as variable rate applications, identification and removal of infected plants, or controlled traffic farming. They can also serve to increase vehicle mobility and safety, providing information about the ground characteristics at various levels.

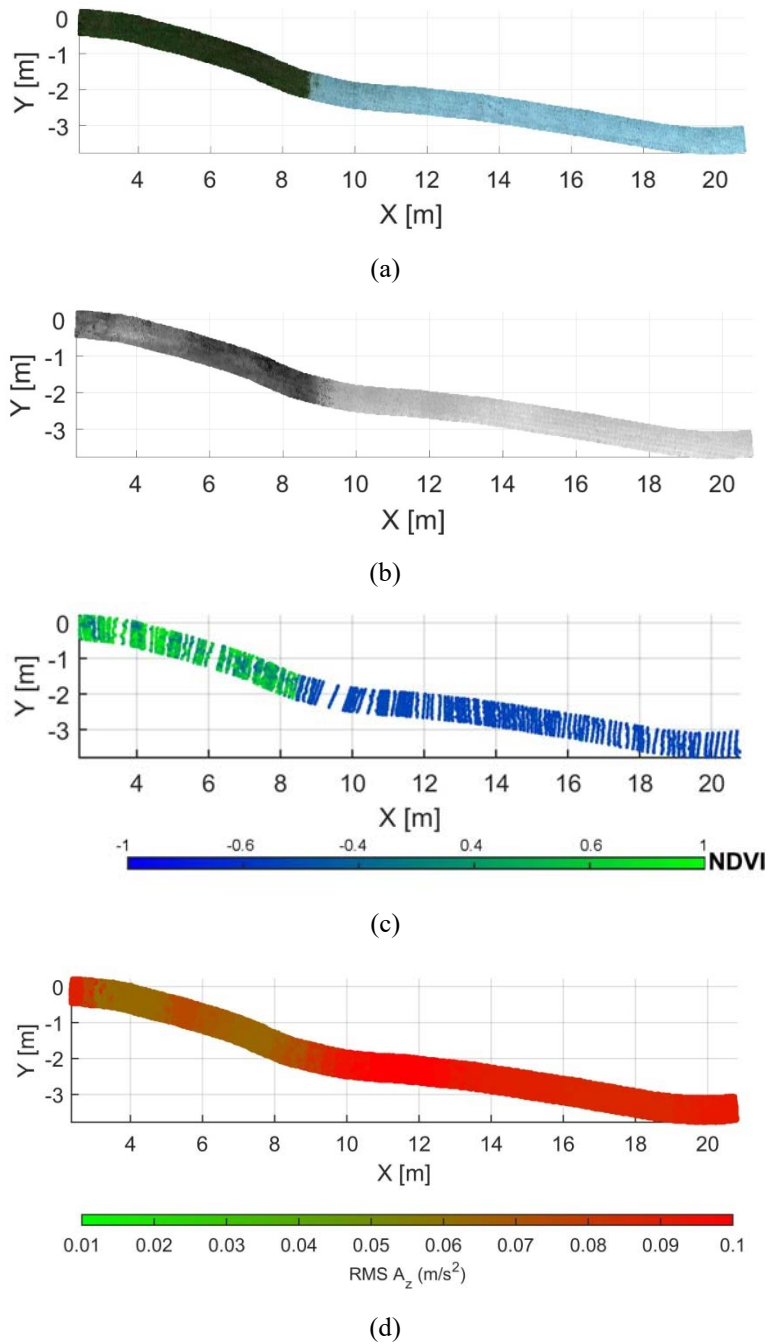

**Figure 14** Multi-layer map obtained for a test on grass and stone-paved road: top view of (a) visual (RGB), (b) thermal, (c) VIS-NIR, and (d) vibration layer.

**Ground change detection**

Stereoscopic, thermal, VIS-NIR and vibration data can be processed to extract informative quantities of the supporting surface in the form of feature vectors capturing the underlying ground properties.

Figure 14 shows, as a case study, the multi-modal map obtained for a test on mixed ground with the vehicle moving first on grass and then on a stone-paved road. For this path, Figure 15 shows the temporal change of the single features extracted from the different sensor types, as explained in section "Description of the feature sets".

All feature graphs show a significant change in the ground properties approximately at 10 s from the start of the vehicle trajectory. This corresponds to the transition from grass to paved ground. In order to automatically detect the change point, the CUSUM test is applied. For simplicity, the result of the change detection algorithm is shown only for the RMS of vertical acceleration in Figure 15(d): when a change is found, a flag is raised (dotted black line). Similar results were obtained also for the other features (Figure 15(a-c)), with a change point instant ranging from 10-12 s. This information can be useful for the robot in many ways, for example, to adjust its control and navigation system.

Future work will include the processing of the maps using supervised or unsupervised classification methods to generate semantic representations of the environment, and for traversability assessment. Slippage detection would add significant information to the feature space. Research efforts will also target the integration of the output maps into farm management information systems (FMIS) to enable map-based control of farming applications. Furthermore, identification and mapping of anomalies, such as stones and weeds, in the data would enable improved targeted farming practices. This would improve the cost-benefit of the sensor suite.

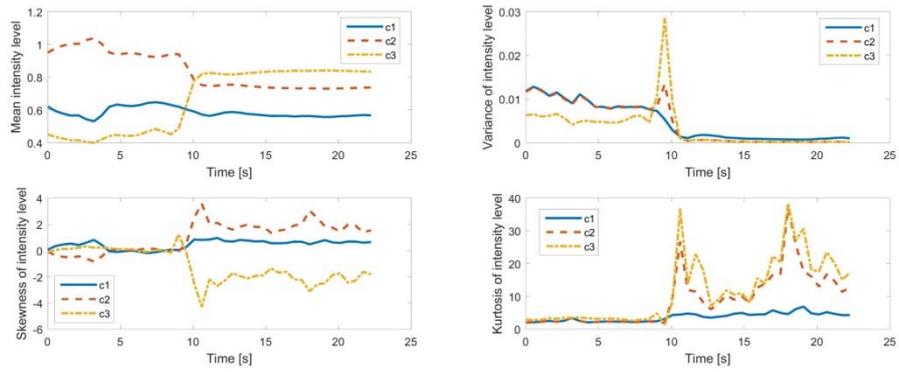

(a)

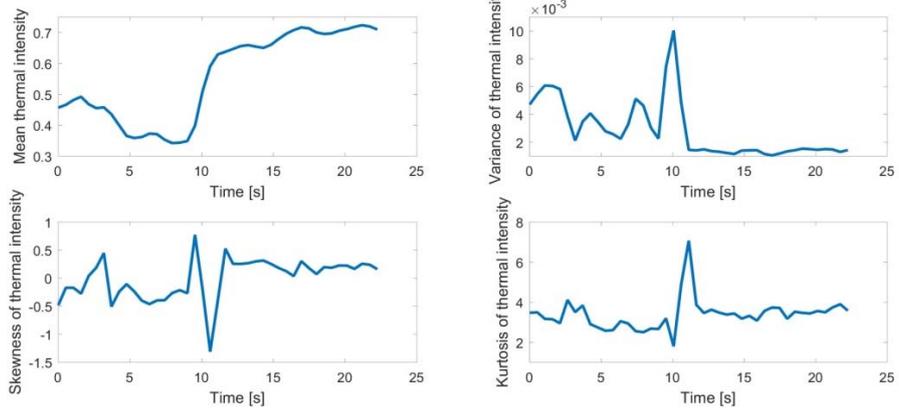

(b)

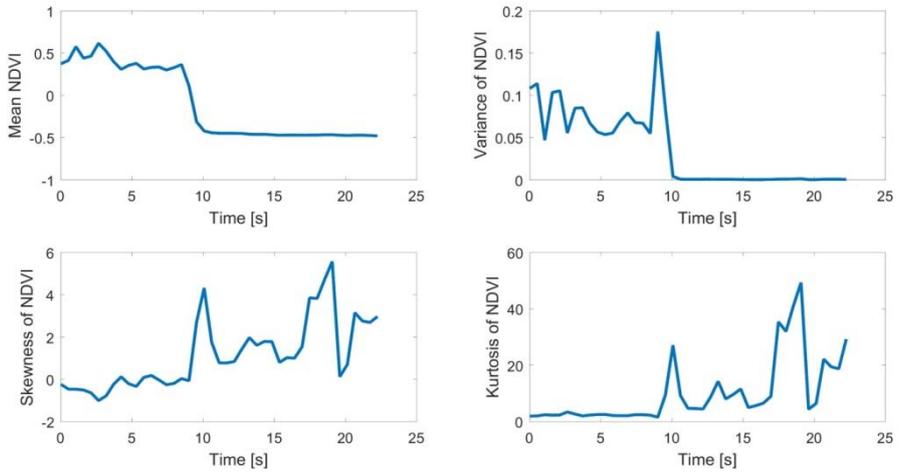

(c)

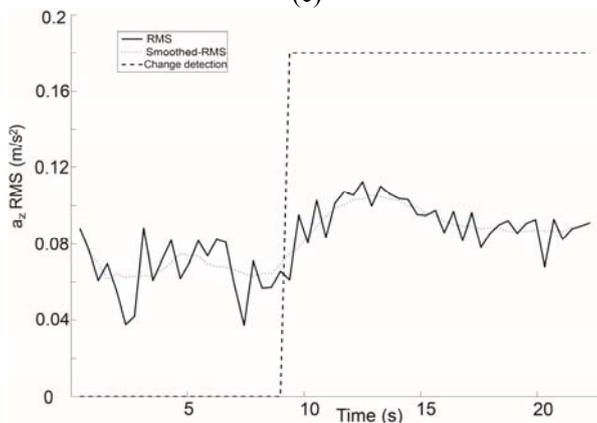

(d)

**Figure 15** Temporal variation of the feature vectors extracted from the different sensor modalities for the test case of Figure 14: (a) visual (color) features, (b) thermal features, (c) NDVI features, (d) RMS of vertical acceleration.

## Conclusions

In this paper, a modular multi-sensor system that incorporates visual, thermal, hyperspectral and inertial sensors for mapping unstructured ground is proposed. It is intended to be used on board an agricultural vehicle to acquire accurate knowledge of the ground characteristics. The design and implementation of the sensor suite on an off-road vehicle was discussed. Then, the integration of all sensor data into 3D ground maps was presented. In addition, a set of features to characterise the ground and detect ground changes was discussed.

Experimental results were included to demonstrate the feasibility of the framework on different ground types. It was shown that the proposed technologies could be adopted to increase the vehicle's ability to perceive and analyse the traversed ground. This could result in higher safety and automation driving levels. For example, information about ground characteristics could be exploited to automatically adjust the vehicle's control and planning strategy to reach more fuel-friendly operation or reduce soil compaction effects.

The proposed multi-sensor integration approach would also allow highly informative soil maps to be built on a narrower scale compared to traditional remote sensing methods. This rich source of information may be made available to the farmer via multimodal interactive maps or farm management information systems, and potentially provide valuable tools to support precision farming applications.

**Acknowledgments**: The financial support of the FP7 ERA-NET ICT-AGRI 2 through the grant Simultaneous Safety and Surveying for Collaborative Agricultural Vehicles (Id. 29839) (S3-CAV) is gratefully acknowledged. The authors would also like to thank the National Research Council (CNR), Italy, for supporting this work under the CNR 2016 Short Term Mobility (STM) program.

**Author Contributions**: Annalisa Milella and Giulio Reina made significant contributions to the conception and design of the research. They mainly dealt with data analysis and interpretation, and writing of the manuscript. Michael Nielsen focused on the development of the multi-sensor system, the experimental activities and data analysis.

# References

Ball, D., Ross, P., English, A., Patten, T., Upcroft, B., Fitch, R., et al. (2015). Robotics for Sustainable Broad-Acre Agriculture Field and Service Robotics. Springer Tracts in Advanced Robotics, 105, 439–453.

Bouguet, J.Y. (2008). Camera calibration toolbox for Matlab, http://www.vision.caltech.edu/bouguetj/calib_doc/. Accessed November 16, 2017.

Brooks, C., Iagnemma, K. (2005). Vibration-based terrain classification for planetary exploration rover. IEEE Transactions on Robotics, 21(6), 1185–1191.

Dong, J., Burnham, J.G., Boots, B., Rains, G., Dellaert, F. (2017). 4D Crop Monitoring: Spatio-Temporal Reconstruction for Agriculture. IEEE International Conference on Robotics and Automation (ICRA), Singapore, 3878-3885.

Dupont, E., Moore, C., Collins, E., Coyle, E. (2008). Frequency response method for terrain classification in autonomous ground vehicles. Autonomous Robots, 24(4), 337–347.

Fehlman, W.L., Hinders, M.K. (2009). Mobile Robot Navigation with Intelligent Infrared Image Interpretation. ISBN 978-1-84882-508-6, DOI 10.1007/978-1-978-1-84882-509-3, London, UK: Springer-Verlag.

Geiger, A., Ziegler, J., Stiller, C. (2011). StereoScan: Dense 3D Reconstruction in Real-time. 2011 IEEE Intelligent Vehicles Symposium, pp 963–968.

Gevers, T., Smeulders, A.W. (1999). Color-based object recognition. Pattern recognition, 32(3), 453–464.

Hirschmuller, H. (2005). Accurate and Efficient Stereo Processing by Semi-Global Matching and Mutual Information. 2005 IEEE Computer Society Conference on Computer Vision and Pattern Recognition (CVPR'05), pp 807-814.

Kragh, M., Jorgensen, R., Pedersen, H. (2015). Object detection and terrain classification in agricultural fields using 3D lidar data. Lecture Notes in Computer Science, 9163, 188–197.

Krebs, A., Pradalier, C., Siegwart, R. (2009). Comparison of Boosting Based Terrain Classification Using Proprioceptive and Exteroceptive Data. In: Khatib O., Kumar V., Pappas G.J. (eds) Experimental Robotics. Springer Tracts in Advanced Robotics, vol 54. Springer, Berlin, Heidelberg, pp 93-102.

Marinello, F., Pezzuolo, A., Gasparini, F., Arvidsson, J., Sartori, L. (2015). Application of the Kinect sensor for dynamic soil surface characterization. Precision Agriculture, 16(6), 601–612.

Milella, A., Reina, G., Underwood, J. (2015). A self-learning framework for statistical ground classification using radar and monocular vision. Journal of Field Robotics, 32(1), 20–41.

Milella, A., Nielsen, M., Reina, G. (2017). Sensing in the visible spectrum and beyond for terrain estimation in precision agriculture. Advances in Animal Biosciences, 8(2), Papers presented at the 11th European Conference on Precision Agriculture (ECPA 2017), Cambridge University Press, pp 423-429.

Mulder, V., de Bruin, S., Schaepman, M., Mayr, T. (2011). The use of remote sensing in soil and terrain mapping - A review. Geoderma, 162, 1–19.

Nieto, J., Monteiro, S., Viejo, D. (2010). 3D geological modelling using laser and hyperspectral data. IEEE International Geoscience and Remote Sensing Symposium (IGARSS). Honolulu, Hawaii, pp 1-7.

Nissimov, S., Goldberger, J., Alchanatis, V. (2015). Obstacle detection in a greenhouse environment using the Kinect sensor. Computers and Electronics in Agriculture, 113, 104–115.

Page, E.S. (1954). Continuous inspection schemes. Biometrika, 41(1-2), 100–115.

Rangel, J., Soldan, S., Kroll, A. (2014). 3D Thermal Imaging: Fusion of Thermography and Depth Cameras. 12th International Conference on Quantitative Infrared Thermography, NDT Open Access Database, pp 1-10.

Reina, G., Milella, A. (2012). Towards Autonomous Agriculture: Automatic Ground Detection Using Trinocular Stereovision. Sensors, 12, 12405–12423.

Reina, G., Milella, A., Galati, R. (2017). Terrain assessment for precision agriculture using vehicle dynamic modelling. Biosystems Engineering, 162, 124–139.

Ross, P., English, A., Ball, D., Upcroft, B., Corke, P. (2015). Online novelty-based visual obstacle detection for field robotics. IEEE International Conference on Robotics and Automation (ICRA), Seattle, Washington, pp 3935–3940.

Rovira-Más, F., Zhang, Q., Reid, J.F. (2008). Stereo vision three-dimensional terrain maps for precision agriculture, Computers and Electronics in Agriculture, 60(2), 133-143.

Underwood, J., Wendel, A., Schofield, B., McMurray, L., Kimber, R. (2017). Efficient in-field plant phenomics for row-crops with an autonomous ground vehicle. Journal of Field Robotics, 34, 1064–1083.

Wu, C., Niu, Z., Tang, Q., Huang, W. (2008). Estimating chlorophyll content from hyperspectral vegetation indices: Modeling and validation. Agricultural and Forest Meteorology, 148 (8-9), 1230–1241.